\documentclass[sigconf]{acmart}

\usepackage{bbm}
\usepackage{subfig}
\usepackage{soul}
\usepackage{color}
\usepackage{amsmath}
\usepackage{amsfonts}
\usepackage{amsthm}
\usepackage{dsfont}

\newtheorem{proposition}{Proposition}

\newtheorem{definition}{Definition}

\AtBeginDocument{%
  \providecommand\BibTeX{{%
    \normalfont B\kern-0.5em{\scshape i\kern-0.25em b}\kern-0.8em\TeX}}}

\copyrightyear{2022}
\acmYear{2022}
\setcopyright{acmcopyright}\acmConference[KDD '22]{Proceedings of the 28th ACM SIGKDD Conference on Knowledge Discovery and Data Mining}{August 14--18, 2022}{Washington, DC, USA}
\acmBooktitle{Proceedings of the 28th ACM SIGKDD Conference on Knowledge Discovery and Data Mining (KDD '22), August 14--18, 2022, Washington, DC, USA}
\acmPrice{15.00}
\acmDOI{10.1145/3534678.3539430}
\acmISBN{978-1-4503-9385-0/22/08}

\begin{document}

\title{Debiasing the Cloze Task in Sequential Recommendation with Bidirectional Transformers}

\author{Khalil Damak}
\authornote{This work was performed while KD and SK were PhD students at University of Louisville. Currently, KD is affiliated with Amazon, and SK is affiliated with Meta.}
\email{khalil.damak@louisville.edu}
\affiliation{%
  \institution{Knowledge Discovery \& Web Mining Lab, Dept. of Computer Science \& Engineering, University of Louisville}
  \streetaddress{Louisville, KY 40292}
  \city{Louisville}
  \state{Kentucky}
  \country{USA}
  \postcode{40208}
}

\author{Sami Khenissi}
\authornotemark[1]
\email{sami.khenissi@louisville.edu}
\affiliation{%
  \institution{Knowledge Discovery \& Web Mining Lab, Dept. of Computer Science \& Engineering, University of Louisville}
  \streetaddress{Louisville, KY 40292}
  \city{Louisville}
  \state{Kentucky}
  \country{USA}
  \postcode{40208}
}

\author{Olfa Nasraoui}
\email{olfa.nasraoui@louisville.edu}
\affiliation{%
  \institution{Knowledge Discovery \& Web Mining Lab, Dept. of Computer Science \& Engineering, University of Louisville}
  \streetaddress{Louisville, KY 40292}
  \city{Louisville}
  \state{Kentucky}
  \country{USA}
  \postcode{40208}
}

\renewcommand{\shortauthors}{Khalil Damak, Sami Khenissi, \& Olfa Nasraoui}

\begin{abstract}
Bidirectional Transformer architectures are state-of-the-art sequential recommendation models that use a bi-directional representation capacity  based on the Cloze task, a.k.a. Masked Language Modeling. The latter aims to predict randomly masked items within the sequence. Because they assume that the true interacted item is the most relevant one, an exposure bias results, where non-interacted items with low exposure propensities are assumed to be irrelevant. The most common approach to mitigating exposure bias in recommendation has been Inverse Propensity Scoring (IPS), which consists of down-weighting the interacted predictions in the loss function in proportion to their propensities of exposure, yielding a theoretically unbiased learning.
In this work, we argue and prove that IPS does not extend to sequential recommendation because it fails to account for the temporal nature of the problem. We then propose a novel propensity scoring mechanism, which can theoretically debias the Cloze task in sequential recommendation. Finally we empirically demonstrate the debiasing capabilities of our proposed approach and its robustness to the severity of exposure bias.
\end{abstract}




\maketitle

\section{Introduction}

Sequential recommendation is a recommendation setting in which the goal is to predict the next best interaction or interactions given a sequence of previous interactions through time \cite{wang2019survey}. Most successful recent work relies on deep learning models including Recurrent Neural Networks (RNNs) \cite{Lipton2015ACR, D14-1179, Hochreiter:1997:LSM:1246443.1246450,hidasi2015session, hidasi2018recurrent}, Convolutional Neural Networks (CNNs) \cite{lecun1999object,tang2018personalized}, and more recently, self-attention modules \cite{vaswani2017attention, devlin2018bert,kang2018self, sun2019bert4rec}.
Recent research has also addressed different biases in recommendation \cite{chen2020bias}. In particular, exposure bias stems from the partial exposure of items to the users \cite{chen2020bias}, making items with relatively low exposure often considered to be irrelevant in building predictive models. Ideally, recommender systems should capture the true relevance of the items to the users, regardless of their propensities of exposure. However, this is far from true on real life recommendation platforms. 
Exposure bias can be mitigated  during the training of recommender systems \cite{chen2020bias}, mainly by making the models aware of the items' exposure propensities. One of the most common approaches consists of building propensity-weighted loss functions that are unbiased estimates of the desirable relevance-based objectives \cite{saito2020unbiased, saito2019unbiased}. This approach, called Inverse Propensity Scoring (IPS), showed success in recommendation settings with user profiles \cite{sun2019debiasing}.
Despite the progress in this area, to the extent of our knowledge, no previous work has addressed the problem of exposure bias in \textit{sequential} recommendation.
In this paper, we  mitigate exposure bias in bi-directional transformer-based recommender systems, which are considered state-of-the-art sequential recommender systems \cite{sun2019bert4rec}, and more specifically, the widely-used BERT4Rec model \cite{sun2019bert4rec}. More broadly however, our work covers any \textit{sequential} recommender system that is trained to optimize the \textit{Cloze} task \cite{taylor1953cloze, devlin2018bert}. Our contributions are summarized as follows: 

\begin{itemize}
    \item We theoretically formulate the problem of exposure bias in the Cloze task, and argue and prove that traditional Inverse Propensity Scoring (IPS) based debiasing frameworks do not extend to sequential recommendation.
    \item We propose an ideal Cloze task loss function that aims to capture the relevance of items within a sequence context.
    \item We propose a novel framework for debiasing the Cloze task in sequential recommendation, called Inverse Temporal Propensity Scoring (ITPS), and use it to propose a novel loss function that produces an unbiased estimator for the ideal Cloze task loss.
    \item We make our implementation available to the public\footnote{\url{https://github.com/KhalilDMK/DebiasedBERT4Rec}}.
    \item We conduct experiments that demonstrate the debiasing capabilities of our ITPS-based estimator, and empirically validate our theoretically proven claims.
\end{itemize}

\section{Background}

Exposure bias occurs when user interactions are dependent upon the exposure of the items. Thus, recommender systems trained on collected data would assume that  interaction represents relevance; and hence, non-interacted items would be considered irrelevant regardless of whether they had a chance to be exposed or not.
Previous work addressing exposure bias varied in whether they treat bias during the \textit{training} or \textit{evaluation} \cite{chen2020bias}. The common approach to mitigating exposure bias in the evaluation of recommender systems relies on incorporating Inverse Propensity Scoring (IPS) in the ranking evaluation metrics. More specifically, items are down-weighted by their popularities in the evaluation metrics \cite{yang2018unbiased}.
On the other hand, a variety of techniques were introduced to mitigate exposure bias in the training phase. Some of these techniques are based on integrating a measure of confidence into the unobserved interactions when considering them as irrelevant. Among these techniques, a few \cite{hu2008collaborative, devooght2015dynamic} considered a uniform weight for all negative items that is lower than one; while others \cite{pan2009mind, pan2008one} utilized user activity, such as the number of interacted items, to weight the negative interactions. Other approaches used item popularity \cite{he2016fast, yu2017selection} and user-item similarity \cite{li2010improving} instead.
Another line of work proposed IPS-based unbiased estimators for the ideal pointwise \cite{saito2020unbiased} and pairwise \cite{saito2019unbiased, damak2021debiased} losses, and estimated the propensity of an interaction using the relative item popularity.
Departing from the previously mentioned methods, some methods proposed new negative sampling processes to mitigate exposure bias during training. This is usually performed by exploiting \textit{side} information such as social network information \cite{chen2019samwalker} or item-based knowledge graphs \cite{wang2020reinforced}.
Another approach consists of integrating the ability to learn the exposure probability within the model by making assumptions on the probability distribution of exposure \cite{liang2016modeling, chen2020fast, chen2019samwalker}.

The above methods share the limitation of recommendation with user profiles, where the goal is to predict items to users regardless of the \textit{temporal} context of the previous interactions. To the extent of our knowledge, no previous work has validated these techniques in sequential recommendation.
Furthermore, only a few studies \cite{zhao2020adversarial, ren2020sequential} have addressed exposure bias in sequential recommendation. However, these approaches treated sequential recommendation in a seq2seq adversarial setting, and use a different formulation of exposure bias which consists of a discrepancy between the training data distribution and the data distribution generated by the model \cite{ranzato2015sequence}, rather than a discrepancy between relevance and interaction.

We address the aforementioned gaps by first studying the limitations of Inverse Propensity Scoring for mitigating exposure bias in sequential recommender systems, and then proposing a debiasing framework that is tailored to sequential recommendation.

\section{Problem Formulation and Motivation}

We start by formulating the sequential recommendation setting before presenting the Cloze task in bidirectional transformer-based models. Next, we discuss the exposure bias problem in the Cloze task, and how the traditional Inverse Propensity Scoring (IPS) framework does not generalize to sequential recommendation.

\subsection{Sequential Recommendation}

Let $S$ be a sequential recommendation dataset comprised of $|S|$ sequences. Each sequence $S_{s}$ is a succession of consecutive item interactions by a user during a certain period of time. An interaction could be defined as a click, rating, review, or consumption, and the time span of the sequence could be short or long. Also, consider a set of items $I$. The sequence $S_{s}$ can be represented by its item interactions, for example $S_{s} = [I_{1}, I_{5}, I_{9}, I_{2}, I_{3}]$. 
We assume that all the sequences are normalized to the same number of time steps $T$ to fit the input requirements of transformer-based models. To do so, sequences that are longer than $T$ time steps are truncated to the most recent $T$ interactions, and sequences that are shorter than $T$ time steps are padded with a padding item $0$ at the beginning. Hence, the dataset $S$ is converted to a matrix $S \in {I \cup \{0\}}^{|S| \times T}$, where element $S_{s, t}$ represents the item, belonging to $I$, in sequence $S_{s}$ at time step $t$. The goal of sequential recommendation is to build a model that is able to accurately predict the next item interaction given a context of previous interactions in a sequence. We represent the trained model by the function $f_{\Omega}$, with parameters $\Omega$, such that $f_{\Omega}: [1, |S|] \times [1, T] \times [1, |I|] \rightarrow \mathbb{R}; (s, t, i) \mapsto f_{\Omega}(S_{s, t}, I_{i})$.
The model $f_{\Omega}$ outputs a prediction of the relevance of item $I_{i}$ for sequence $S_{s, t}$ at time step $t$.
More specifically, in our work, $f_{\Omega}$ is the bi-directional transformer-based model BERT4Rec \cite{sun2019bert4rec}. Because the use of Transformers has become common, and because our focus is on debiasing the Cloze task rather than the model itself, we omit an exhaustive background description of transformers, and the BERT4Rec model architecture. Instead, we refer the reader to \cite{sun2019bert4rec}.
That said, we note that all the findings described in this paper are model-agnostic, as long as the model is trained for the Cloze task, and is capable of modeling sequential data.

\subsection{The Cloze Task in Sequential Recommendation}

The Cloze task \cite{taylor1953cloze} consists of randomly masking a percentage $\rho$ of the tokens, in our case items in the sequence, and training the machine learning model to predict those masked tokens. This approach, also called ``Masked Language Modeling" (MLM) \cite{devlin2018bert}, allows for learning a bidirectional context in the training sequence without any information leakage \cite{sun2019bert4rec} from the future. 
This ability of modeling a bidirectional context through the Cloze objective is what gives BERT4Rec its prediction power compared to other models, such as uni-directional self-attention based recommender systems \cite{vaswani2017attention}.
Consider a training dataset $S^{m} \in {I \cup \{0, \langle mask \rangle\}}^{|S| \times T}$. $S^{m}$ is a masked version of the ground truth dataset $S$ where a fraction $\rho$ of the items is replaced with the token $\langle mask \rangle$ in each sequence.
The goal of the Cloze task is to train the hypothesis $f_{\Omega}$ to reconstruct the ground truth dataset $S$ from the masked training dataset $S^{m}$. Hence, the loss function associated with the Cloze task is defined as the negative log-likelihood of the predicted probability of correctly predicting the masked tokens, which we formulate as follows:

\begin{definition}[Cloze Task Loss Function]\label{def:colze_loss}

\begin{equation}\label{eq:cloze_loss}
\begin{aligned}
    L_{Cloze} = &\frac{-1}{|S| |I| T} \sum_{s = 1}^{|S|} \sum_{t = 1}^{T} \sum_{i = 1}^{|I|} \mathds{1}_{\{ S^{m}_{s, t} = \langle mask \rangle \}} Y_{S_{s}, I_{i}, t}\\
    &\times log \; softmax(f_{\Omega}(S^{m}_{s, t}, I_{i}))
\end{aligned}
\end{equation}

\end{definition}

\noindent
where $softmax(f_{\Omega}(S^{m}_{s, t}, I_{i})) = \frac{e^{f_{\Omega}(S^{m}_{s, t}, I_{i})}}{\sum_{k = 1}^{|I|} e^{f_{\Omega}(S^{m}_{s, t}, I_{k})}}$ approximates the predicted probability $P(S_{s, t} = I_{i} | S^{m}_{s})$ of the ground truth item in sequence $S_{s}$ at time $t$ being $I_{i}$ given the masked sequence $S^{m}_{s}$. $Y_{S_{s}, I_{i}, t}$ is a binary random variable that equals 1 when $I_{i} \in I$ is interacted with in sequence $S_{s} \in S$ at time step $t \in [1, T]$, and 0 otherwise.


\subsection{Exposure Bias in the Cloze Task}

The Cloze loss function, in Definition \ref{def:colze_loss}, considers the interacted ground truth item $S_{s, t}$ as the desirable and relevant target item for the input $S^{m}_{s, t}$. However, as shown in \cite{saito2020unbiased, saito2019unbiased, schnabel2016recommendations}, interaction does not necessarily signify relevance. In other words, an item could be interacted because it was the most relevant item among the items that the user was exposed to within the item sequence at the corresponding time step. Moreover, non-interacted items could be relevant to some extent, and it could be that the user did not interact with them because they were not exposed to the user. It is this estimation of the relevance of an item with the interaction that engenders the exposure bias.
Hence, we can define the ideal Cloze task loss by replacing the interaction random variable $Y_{S_{s}, I_{i}, t}$ by the relevance of the item that the user chose to interact with in sequence $S_{s}$ at time step $t$, assuming that the user is aware of all items. The awareness of the user of all items completely eliminates the exposure bias because it infers that all items were exposed to the user. Moreover, weighting the interaction by the relevance allows the loss to capture the true relevance of the item.
Hence, we consider a Bernoulli random variable $R_{S_{s}, I_{i}, t} \sim Ber(\gamma_{S_{s}, I_{i}, t})$, where $\gamma_{S_{s}, I_{i}, t} = P(R_{S_{s}, I_{i}, t} = 1)$ represents the probability of item $I_{i}$ being relevant in sequence $S_{s}$ at time step $t$ (i.e., $R_{S_{s}, I_{i}, t}$ equals 1).
Moreover, we define a Choice random variable that simulates the user behaviour when choosing to interact with item $I_{i}$ within sequence $S_{s}$ at time step $t$. We assume that this choice is contingent upon its relevance compared to all the other items given the sequence context. Hence, we can model the Choice random variable $C_{S_{s}, I_{i}, t}$ by a Categorical (Generalized Bernoulli) distribution as follows:

\begin{equation}
    C_{S_{s}, t} \sim Cat(|I|, [\gamma_{S_{s}, I_{1}, t}, .., \gamma_{S_{s}, I_{|I|}, t}]).
\end{equation}

The outcome of the random variable is a vector of $|I|$ zeroes except for a 1 for the item the user chooses to interact with. This means that the user chooses one of the $|I|$ items based on their relevance to the context $S_{s, t}$. We denote the outcome of $C_{S_{s}, t}$ for item $I_{i}$ by $C_{S_{s}, I_{i}, t}$ and define the ideal Cloze task loss as follows:

\begin{definition}[Ideal Cloze Task Loss Function]

\begin{equation}\label{eq:ideal_loss}
\begin{aligned}
    L_{Cloze}^{ideal} = &\frac{-1}{|S| |I| T} \sum_{s = 1}^{|S|} \sum_{t = 1}^{T} \sum_{i = 1}^{|I|} \mathds{1}_{\{ S^{m}_{s, t} = \langle mask \rangle \}} C_{S_{s}, I_{i}, t}\\ 
    &\times \gamma_{S_{s}, I_{i}, t} log \; softmax(f_{\Omega}(S^{m}_{s, t}, I_{i})).
\end{aligned}
\end{equation}

\end{definition}

The discrepancy between the interaction random variable $Y_{S_{s}, I_{i}, t}$ and the product $C_{S_{s}, I_{i}, t} \; \gamma_{S_{s}, I_{i}, t}$ causes the Cloze task loss to be biased against the ideal loss, as stated in the following Proposition:

\begin{proposition}[Exposure Bias of the Cloze Task Loss Function]\label{prop:bias_cloze_loss}
The Cloze task loss function is biased against the ideal Cloze task loss, such that $\mathbb{E}[L_{Cloze}] \neq L_{Cloze}^{ideal}.$
See Appendix \ref{app:bias_cloze_loss} for proof.

\end{proposition}

\subsection{Inverse Propensity Scoring in the Cloze Task and Its Limitations}

The common solution to debiasing a maximum likelihood-based loss function for recommendation is Inverse Propensity Scoring (IPS) where an IPS-based estimator of the ideal pointwise loss is obtained by weighting every item prediction for a user by the reciprocal of its exposure propensity for that user \cite{saito2020unbiased}. The IPS framework is suitable for debiasing loss functions for recommendation with user profiles. However, we argue that it does not extend to sequential recommendation for the following two reasons:

\textbf{(1) Inadequacy of the interaction random variable representation:}
The IPS-based framework for recommendation with user profiles \cite{saito2020unbiased} models the interaction random variable $Y_{u, i}$, that represents whether user $u$ interacted with item $i$, by the product of the relevance and the exposure of the item to the user. The framework relies on two random variables, $O_{u, i} \sim Ber(\theta_{u, i})$ and $R_{u, i} \sim Ber(\gamma_{u, i})$, of exposure and relevance respectively, and models the interaction using $Y_{u, i} = O_{u, i} R_{u, i}$.
This means that an item is interacted with by a user if and only if it is both observed by, and relevant to the user. If we extend this modeling of the interaction to sequential recommendation by mapping users to sequences and introducing the temporal component, we would obtain for a sequence $S_{s}$, an item $I_{i}$ and a time step $t$: $Y_{S_{s}, I_{i}, t} = O_{S_{s}, I_{i}, t} R_{S_{s}, I_{i}, t}$, where $R_{S_{s}, I_{i}, t}$ is the relevance random variable and $O_{S_{s}, I_{i}, t}$ is a Bernoulli exposure random variable that takes value 1 if item $I_{i}$ was exposed in sequence $S_{s}$ at time step $t$, such that $O_{S_{s}, I_{i}, t} \sim Ber(\theta_{S_{s}, I_{i}, t})$. $\theta$ is the probability of exposure such that $\theta_{S_{s}, I_{i}, t} = P(O_{S_{s}, I_{i}, t} = 1)$.
This modeling of the interaction random variable is inadequate for sequential recommendation. In fact, in traditional recommendation, it is safe to assume that any item that is exposed and relevant  to a user is interacted. However, when introducing the temporal component into the equation, the assumption does not hold anymore. This is because a user can only interact with one item at a time. Multiple items can be relevant for the same sequence at the same time step, but only one of them can be interacted with. For this reason, the IPS-based framework for recommendation with user profiles does not extend to sequential recommendation.

\textbf{(2) Ignoring the temporal component:}
The IPS estimator for the ideal pointwise loss function down-weights every interaction $Y_{u, i}$ by the propensity of exposure of item $i$ to user $u$, $\theta_{u, i}$. In order to define an IPS-based Cloze loss for sequential recommendation, we assimilate the users to sequences and consider the propensity of exposure of an item $I_{i}$ in a sequence $S_{s}$ as $\theta_{S_{s}, I_{i}} = P(O_{S_{s}, I_{i}} = 1)$, where $O_{S_{s}, I_{i}} \sim Ber(\theta_{S_{s}, I_{i}})$ is a Bernoulli random variable that takes the value $1$ when item $I_{i}$ is exposed in sequence $S_{s}$. We define the IPS-based Cloze loss as follows:

\begin{definition}[Inverse Propensity Scoring-based Cloze Loss Function]

\begin{equation}\label{eq:IPS_cloze_loss}
\begin{aligned}
    L_{Cloze}^{IPS} = &\frac{-1}{|S| |I| T} \sum_{s = 1}^{|S|} \sum_{t = 1}^{T} \sum_{i = 1}^{|I|} \mathds{1}_{\{ S^{m}_{s, t} = \langle mask \rangle \}} \frac{Y_{S_{s}, I_{i}, t}}{\theta_{S_{s}, I_{i}}}\\ 
    &\times log \; softmax(f_{\Omega}(S^{m}_{s, t}, I_{i})).
\end{aligned}
\end{equation}

\end{definition}

The IPS-based Cloze loss function can only be completely unbiased if the propensity of every item $I_{i}$ in every sequence $S_{s}$ at time step $t$, $\theta_{S_{s}, I_{i}, t}$, is equal to the ``static" propensity, $\theta_{S_{s},I_{i}}$, of item $I_{i}$ in sequence $S_{s}$. We state this in the following proposition:

\begin{proposition}[Unbiasedness condition of the IPS-based Cloze loss function]\label{prop:unbiasedness_condition}

\begin{equation}\label{eq:unbiasedness_condition_real}
    \mathbb{E}[L_{Cloze}^{IPS}] = L_{Cloze}^{ideal} \Leftrightarrow \theta_{S_{s}, I_{i}, t} = \theta_{S_{s}, I_{i}}, \forall (S_{s}, I_{i}, t) \in S \times I \times [1..T].
\end{equation}

\end{proposition}

\begin{figure}
\begin{center}
\includegraphics[width=.48\linewidth]{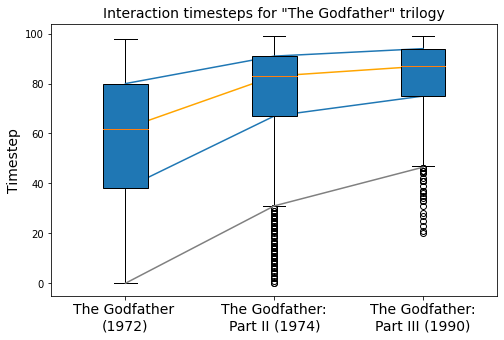}
\includegraphics[width=.48\linewidth]{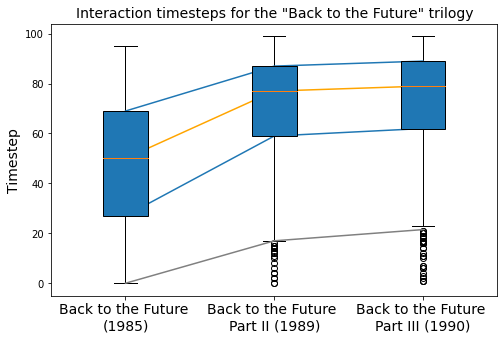}
\end{center}
\caption{Boxplots of the interaction timesteps for "The Godfather" and "Back to the Future" trilogies. The interaction distributions vary through time, meaning that the exposure propensities must not be considered static. \label{fig:interaction_timesteps_boxplots}}
\end{figure}

The proof is in Appendix \ref{app:proof_unbiasedness_condition}.
This unbiasedness condition of the IPS estimator is unlikely and hard to satisfy as the propensities of exposure tend to vary with the temporal context. We demonstrate this in Figure \ref{fig:interaction_timesteps_boxplots} where we show boxplots of the interaction time steps for two movie trilogies in the Movielens 1M dataset \cite{10.1145/2827872}. The boxplots show that there are movies that tend to be watched later than others in the sequence; for instance, sequels tend to be watched after the original movies. We chose movies that are older than the dataset to ensure that the differences in observation time are not related to the release dates of the movies, but rather to the temporal context within the trilogies. Hence, given that the interaction distribution tends to vary with time, it is safe to assume that the exposure propensities also vary with time. Thus, in contrast to the IPS framework, they should not be considered static in sequential recommendation.
The latter observation additionally shows how the IPS framework does not extend to sequential recommendation. This consequently calls for proposing a new framework that is specifically tailored for debiasing the Cloze task in sequential recommendation, which is the subject of the next section.

\section{Inverse Temporal Propensity Scoring for an Unbiased Cloze Task}\label{sec:itps_section}

The Inverse Propensity Scoring technique fails to capture the temporal component of the sequential recommendation setting, and hence fails to provide an unbiased estimation of the ideal Cloze task loss. We propose a debiasing framework that is tailored to the Cloze task in sequential recommendation, and that we call \textbf{Inverse Temporal Propensity Scoring (ITPS)}. In ITPS, we address the two main limitations of IPS that prevent it from generalizing to sequential recommendation.
First, to address the issue of the inadequacy of the interaction random variable representation, we include the outcome of the Choice random variable for item $I_{i}$ in the interaction model for the following formulation:

\begin{definition}[Interaction Random Variable Representation in the ITPS Framework]\label{def:interaction_probabilistic_model}
\begin{equation}\label{eq:ineraction_model}
    Y_{S_{s}, I_{i}, t} = C_{S_{s}, I_{i}, t} \; O_{S_{s}, I_{i}, t} \; R_{S_{s}, I_{i}, t}
\end{equation}
\end{definition}

The latter formulation of the interaction allows for only one item to be interacted within a sequence at a given time step, which is adequate for sequential recommendation. Now, an item $I_{i}$ is interacted by a user ($Y_{S_{s}, I_{i}, t} = 1$) in a sequence $S_{s}$ at time step $t$ if and only if the item is exposed ($O_{S_{s}, I_{i}, t} = 1$), relevant ($R_{S_{s}, I_{i}, t} = 1$), and chosen by the user based on its relevance ($C_{S_{s}, I_{i}, t} = 1$).
Finally, to account for the temporal component in sequential recommendation in ITPS, we weight the prediction of every item $I_{i}$ in every sequence $S_{s}$ at every time step $t$ by the temporal propensity $\theta_{S_{s}, I_{i}, t}$, as opposed to the static propensity $\theta_{S_{s}, I_{i}}$ of IPS. Thus, we define the ITPS-based Cloze task loss function as follows:

\begin{definition}[Inverse Temporal Propensity Scoring-based Cloze Loss Function]

\begin{equation}\label{eq:ITPS_cloze_loss}
\begin{aligned}
    L_{Cloze}^{ITPS} = &\frac{-1}{|S| |I| T} \sum_{s = 1}^{|S|} \sum_{t = 1}^{T} \sum_{i = 1}^{|I|} \mathds{1}_{\{ S^{m}_{s, t} = \langle mask \rangle \}} \frac{Y_{S_{s}, I_{i}, t}}{\theta_{S_{s}, I_{i}, t}}\\
    &\times log \; softmax(f_{\Omega}(S^{m}_{s, t}, I_{i}))
\end{aligned}
\end{equation}

\end{definition}

This new ITPS-based loss is an unbiased estimator of the ideal Cloze task loss, as stated in the following proposition:

\begin{proposition}\label{prop:itps_unbiased}
The ITPS-based Cloze task loss is unbiased for the ideal Cloze task loss, meaning that $\mathbb{E}[L_{Cloze}^{ITPS}] = L_{Cloze}^{ideal}.$

\end{proposition}

The proof is in Appendix \ref{app:proof_itps_unbiased}.



\section{Experimental Evaluation}

We perform experiments to assess the validity of our theoretical claims of unbiasedness and the applicability of our approach in real recommendation settings. We use  semi-synthetic and real world datasets.  The semi-synthetic data, used in Section \ref{sec:semi_synthetic_experiments}, provides a full visibility of the data properties, allowing us to evaluate the debiasing capabilities of our proposed approach. Moreover, it allows us to control the data properties in order to evaluate the robustness of our approach to varying bias levels. The real datasets, used in Section \ref{sec:real_experiments}, allow us to evaluate the applicability of our approach in real recommendation settings. Additionally, we simulate a feedback loop to evaluate the longitudinal effects of the proposed debiasing.

\subsection{Experiments on Semi-Synthetic Data}\label{sec:semi_synthetic_experiments}

We perform experiments to answer three research questions:

\textbf{RQ1: }How well does the proposed ITPS estimator capture the true relevance?

\textbf{RQ2: }How robust is the proposed ITPS estimator to increasing levels of exposure bias?

\textbf{RQ3: }How important is an unbiased evaluation in assessing exposure debiasing?


\subsubsection{Data} \label{sec:semi_synthetic_data_creation}

\begin{table}
\caption{Statistics of the real (ml-100k) and semi-synthetic (ss-ml-100k) Movielens 100K datasets. \label{tab:data_statistics} }
\begin{center}
{
\resizebox{\columnwidth}{!}{\begin{tabular}{ l | c c c c c}
\hline
Dataset & \# sequences & \# items & \# ratings & Avg. length & Sparsity\\ \hline \hline
ml-100k & 943 & 1,349 & 99,287 & 105.28 & 92.19\%\\
ss-ml-100k & 943 & 229 & 94,104 & 99.79 & 56.42\%\\
\hline
\end{tabular}}
}
\end{center}
\end{table}

Semi-synthetic experiments are necessary due to the unavailability of any open or public unbiased sequential recommendation dataset. In fact, only an exposure-unbiased testing dataset would allow us to truly compare the debiasing capabilities of the different approaches - a claim that we validate in RQ3. 
We use the Movielens 100K (ml-100k)\footnote{https://grouplens.org/datasets/movielens/100k/} dataset because it is a benchmark dataset that can be used for sequential recommendation since it includes interaction timestamps. This data is described in the first row of Table~\ref{tab:data_statistics}. The choice of this dataset is justified due to its relatively low number of sequences (users) and items, compared to other sequential datasets.
In fact, our first task is to generate all data properties, including relevance, exposure, and interaction for all sequence, item and timestep tuples; a task that is resource-expensive, especially in memory requirements. Considering a dataset with $|S|$ sequences, $|I|$ items and $T$ time steps, the number of parameters that need to be predicted and kept into memory for each controlled property is $|S| \times |I| \times T$. Hence, given the ml-100k dataset statistics, we would be predicting over 127 Million values for every property. For this reason, using other benchmark datasets with tens of thousands of sequences or items, is simply prohibitive with our current resources. Moreover, similar conclusions could be drawn regardless of the dataset, assuming a high reconstruction quality.
Our goal is to use the available ratings to infer all the data properties, namely the relevance, exposure, and interaction of all items $I_{i} \in I$, in all sequences $S_{s} \in S$, and at all time steps $t \in [1, T]$. This is done in the following steps:

    \textbf{(1)} We normalize the dataset to $T=100$ time steps.
    
    \textbf{(2)} We train a Tensor Factorization (TF) model \cite{zhao2021tbtf, adomavicius2005incorporating} on the available (sequence, item, timestep, rating) tuples to reconstruct the missing ratings. 
    We train the model on the Mean Squared Error (MSE) loss for rating prediction. Finally, we use the trained TF model to reconstruct the rating tensor by predicting the missing ratings. Given that the rating represents an explicit measure of satisfaction of a user with an item, we can approximate the probability of relevance of an item $I_{i}$ in a sequence $S_{s}$ at a time step $t$ by normalizing the predicted rating with the sigmoid function as follows: $\gamma_{S_{s}, I_{i}, t} \approx \sigma(\hat{r}_{s, i, t})$. Here, $\hat{r}_{s, i, t} = \sum_{k=1}^{d} P_{s, k} Q_{i, k} W_{t, k}$ is the predicted rating, where $P$, $Q$, and $W$ are respectively the sequence, item, and time latent factor matrices, which all have $d$ latent features.
    
    \textbf{(3)} We train another Tensor Factorization model to predict the probabilities of exposure. We convert every rating in the dataset to a positive exposure, and sample a portion of non-interacted tuples as negative exposures. We assume that an item has a higher probability of not being exposed than of being exposed, which is a realistic assumption given the abundance of items in recommendation platforms. Thus, we sample 3 negative exposure tuples for every positive exposure tuple. 
    We train the TF model using the Binary Cross Entropy loss for exposure classification.
    Similarly to step (2), we approximate the propensity of exposure of an item $I_{i}$ in a sequence $S_{s}$ at a time step $t$ by the predicted exposure as follows: $\theta_{S_{s}, I_{i}, t} \approx \hat{o}_{s, i, t}$. Here, $\hat{o}_{s, i, t}$ is the predicted exposure probability of item $i$ in sequence $s$ at time step $t$, obtained by: $\hat{o}_{s, i, t} = \sigma(\sum_{k=1}^{d} P_{s, k} Q_{i, k} W_{t, k})$.
    
    
    
    \textbf{(4)} Following \cite{saito2020unbiased}, we introduce a hyperparameter $p$ that controls the skewness of the exposure distribution, and hence the level of exposure bias, as follows:
    
    \begin{equation}\label{eq:propensity_power}
        \theta_{S_{s}, I_{i}, t} \approx \hat{o}_{s, i, t}^{p}.
    \end{equation}
    
    The higher the value of $p$, the higher the level of exposure bias introduced. We will control the value of $p$ to study RQ2.
    
    \textbf{(5)} We generate the interaction random variable for every sequence $S_{s}$, item $I_{i}$, and timestep $t$ combination by following the probabilistic model presented in Equation~\ref{eq:ineraction_model}, such that:
    
    \begin{align}
        O_{S_{s}, I_{i}, t} \sim Ber(\theta_{S_{s}, I_{i}, t})\\
        R_{S_{s}, I_{i}, t} \sim Ber(\gamma_{S_{s}, I_{i}, t})\\
        C_{S_{s}, I_{i}, t} \sim Cat(|I|, [\gamma_{S_{s}, I_{1}, t}, .., \gamma_{S_{s}, I_{|I|}, t}])\\
        Y_{S_{s}, I_{i}, t} = C_{S_{s}, I_{i}, t} \; O_{S_{s}, I_{i}, t} \; R_{S_{s}, I_{i}, t}.
    \end{align}
    
    In our experiments, we obtain $C_{S_{s}, I_{i}, t}$ by considering a rational user interacting with the exposed item ($O_{S_{s}, I_{i}, t} = 1$) with highest relevance $\gamma_{S_{s}, I_{i}, t}$.
    
    \textbf{(6)} Finally, we filter the interacted instances to construct the semi-synthetic sequential dataset. The statistics of a sample generated semi-synthetic dataset are presented in the second row of Table~\ref{tab:data_statistics}.


\subsubsection{Evaluation Process}\label{sec:evaluation_process}

Our estimators should be evaluated in terms of their capacity to capture the true relevance of the test interactions.
However, our sequence interactions are obtained with the interaction probabilistic model in Equation~\ref{eq:ineraction_model}, which requires all interactions to be exposed. Hence, sampling the test and validation interactions from the semi-synthetic sequences would not allow for an evaluation in terms of the true relevance. This is because the most relevant items are not necessarily exposed to the user. We cope with this issue using the following evaluation process:
We start by splitting the data into training, validation and test sets by considering the last item interaction in each sequence for testing and the second to last for validation. Then, we replace every item interaction in the validation and test sets by the item $I_{i}$ with the highest relevance $\gamma_{S_{s}, I_{i}, t}$ in the corresponding sequence $S_{s}$ and at the corresponding timestep $t$. This way, the model is evaluated on its ability to predict the most relevant item, which translates to its ability to capture the true relevance of the items.
This being done, we compare the ranking of the test and validation instances to 100 randomly sampled items. Note that negative sampling does not introduce any bias because, regardless of their exposure, all the negative items are less relevant than the test and validation items.
Thus, our evaluation process is unbiased and evaluates the models in terms of their capacity to capture the true relevance of the items.
We use Normalized Discounted Cumulative Gain ($NDCG@k$) and Recall ($R@k$) for the ranking evaluation.



\subsubsection{Models Compared}\label{sec:models_compared}

We compare the following models:
\begin{itemize}
    \item \textbf{BERT4Rec: }This is the original BERT4Rec model trained to optimize the Cloze task loss in Equation~\ref{eq:cloze_loss}. It relies solely on the interaction information and does not incorporate any exposure debiasing.
    \item \textbf{IPS-BERT4Rec: }This is the BERT4Rec model trained with the IPS-based Cloze loss function in Equation~\ref{eq:IPS_cloze_loss}. We estimate the ``static" exposure propensities by averaging the temporal exposures, such that $\theta_{S_{s}, I_{i}} = \frac{1}{T} \sum_{t=1}^{T} \theta_{S_{s}, I_{i}, t}, \forall (S_{s}, I_{i}) \in S \times I$.
    \item \textbf{ITPS-BERT4Rec: }This is the BERT4Rec model trained with our ITPS-based Cloze task loss in Equation~\ref{eq:ITPS_cloze_loss}. The loss relies on the temporal propensities $\theta_{S_{s}, I_{i}, t}$ to provide an unbiased estimation of the ideal Cloze task loss.
    \item \textbf{Oracle: }This is the BERT4Rec model trained with the ideal Cloze task loss in Equation~\ref{eq:ideal_loss}. The loss has access to the true relevance of the items $\gamma_{S_{s}, I_{i}, t}$ in the training, and hence, is able to provide a completely unbiased representation of the user preferences. Hence, this model provides an upper bound on capturing the true relevance.
\end{itemize}

Because the goal of the experiments is to assess the impact of the different debiasing frameworks, we leave the comparison to additional baselines for future work.

\subsubsection{Hyperparameter Tuning}\label{sec:tuning_semi_synthetic}

We tune all the models presented in Section~\ref{sec:models_compared}, along with the Tensor Factorization models presented in steps 2 and 3 of Section~\ref{sec:semi_synthetic_data_creation} as described below.

\textbf{Tuning the BERT4Rec models:}
Using random search, we tune the number of hidden units within the set \{8, 16, 32, 64\}, the number of transformer blocks within \{1, 2\}, the number of attention heads within \{1, 2\}, the batch size within \{8, 16, 32\}, the dropout rate within \{0, 0.1, 0.2, 0.4\}, and finally, the masking probability $\rho$ of the Cloze task within \{0.1, 0.15, 0.2, 0.4, 0.6\}. We try 30 random combinations, and compare the average $NDCG@10$ results over 3 replicates on the validation set. 

\textbf{Tuning the Tensor Factorization models:}
We randomly split the data into training, validation and test sets with the respective ratios 80\%, 10\% and 10\%. We adopt a grid search by trying all combinations of number of latent features within \{50, 100, 200\}, and batch size within \{64, 128, 256\}. We replicate every experiment 3 times and compare the average performances on the validation set. The rating-based TF model from step 2 is tuned in terms of Mean Squared Error (MSE) for rating prediction, while the exposure-based TF-model from step 3 is tuned in terms of Area Under the ROC Curve (AUC) for exposure classification.


\subsubsection{RQ1: How well does the proposed ITPS estimator capture the true relevance?}\label{sec:rq1_results}

\begin{table}
\caption{Model comparison in terms of capturing the true relevance: Average Recall@k and NDCG@k results over 5 replicates. The best results are in \textbf{bold} and second to best results are \underline{underlined}. A value with * is significantly higher than the next best value (p-value $< 0.05$).\label{tab:relevance_results} }
\begin{center}
{
\resizebox{\columnwidth}{!}{\begin{tabular}{ l | c c c c}
\hline
Model & R@10 & NDCG@10 & R@5 & NDCG@5\\ \hline \hline
BERT4Rec & 0.7992 & 0.6065 & 0.6917 & 0.5716\\
IPS-BERT4Rec & 0.7890 & 0.5961 & 0.6868 & 0.5628\\
ITPS-BERT4Rec & \underline{0.8027}* & \underline{0.6110}* & \underline{0.6997}* & \underline{0.5777}*\\
Oracle & \textbf{0.8218}* & \textbf{0.6247}* & \textbf{0.7083}* & \textbf{0.5880}*\\
\hline
\end{tabular}}
}
\end{center}
\end{table}

To answer this research question, we evaluate the models in terms of their capacity to capture the true relevance using the evaluation process described in Section~\ref{sec:evaluation_process}. We summarize the  results in Table~\ref{tab:relevance_results}.
The best performer on all metrics is the Oracle model, owing to its explicit optimization using the relevance levels. 
The ITPS-BERT4Rec model was second-to-best in all configurations, outperforming the naive BERT4Rec and IPS-BERT4Rec. These findings demonstrate the power of the ITPS debiasing framework and validate the theoretical claims of exposure debiasing of the proposed estimator. Finally and interestigly, IPS-BERT4Rec performed worse than the naive BERT4Rec. This is probably due to the fact that it is trained on estimated static propensities, obtained by averaging the temporal propensities, rather than true propensities.

\subsubsection{RQ2: How robust is the proposed ITPS estimator to increasing levels of exposure bias?}

\begin{figure}
\begin{center}
\includegraphics[width=.4\linewidth]{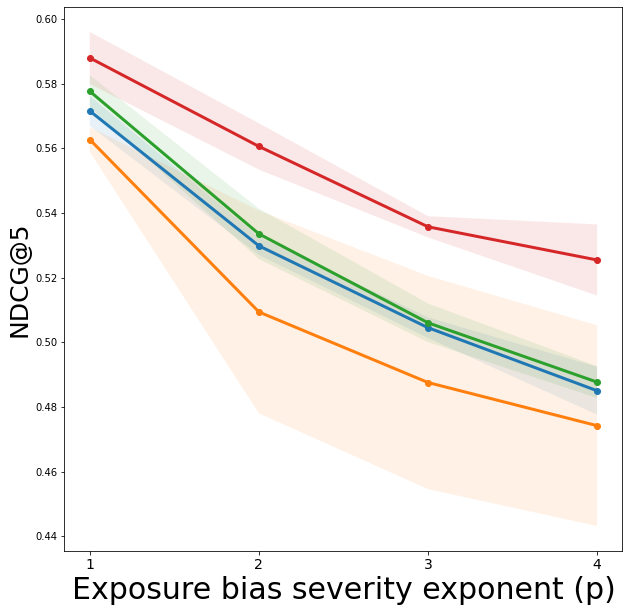}
\includegraphics[width=.4\linewidth]{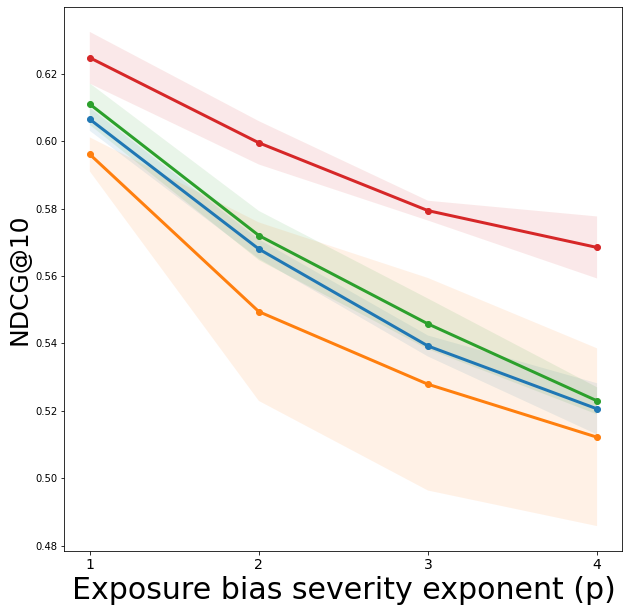}
\includegraphics[width=.4\linewidth]{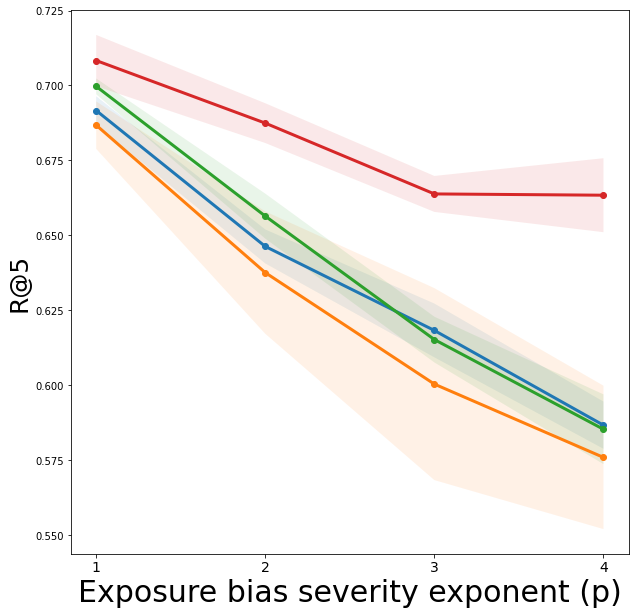}
\includegraphics[width=.4\linewidth]{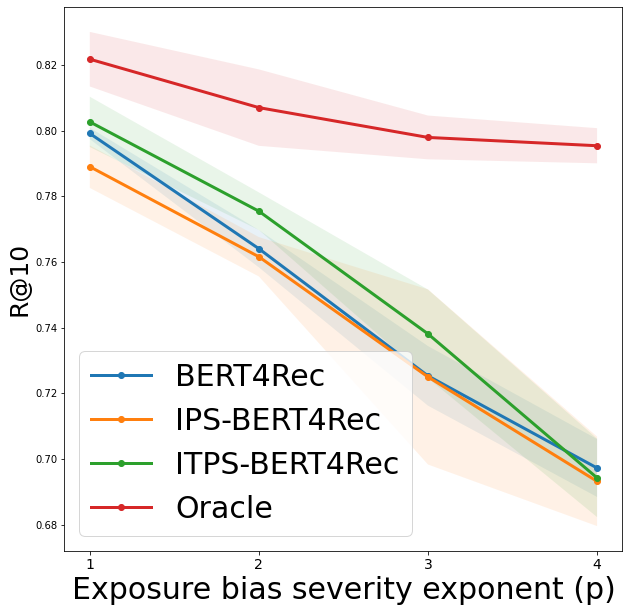}
\end{center}
\caption{Robustness of the ranking performance of the different models to increasing levels of exposure bias. All the values are averages over 5 replicates and the 90\% confidence intervals are highlighted. ITPS-BERT4Rec was the best in withstanding increasing levels of exposure bias overall. \label{fig:evolution_bias_levels}}
\end{figure}

To answer this research question, we train and evaluate the models on semi-synthetic datasets generated with increasing levels of exposure bias. The level of exposure bias is controlled by the power $p$ that governs the propensities $\theta_{S_{s}, I_{i}, t}$ in Equation~\ref{eq:propensity_power}. We increase $p$ from 1 to 4 with an increment of 1, where the higher the value of $p$, the stronger the exposure bias introduced in the data, and show the evolution of the ranking metrics in Figure~\ref{fig:evolution_bias_levels}.
All the models' performances decrease with increasing levels of exposure bias, however with different slopes. The IPS-BERT4Rec model shows the worst performance in handling increasing exposure bias. Its performance quickly degrades starting from $p=2$. This shows the inability of the IPS framework to mitigate exposure bias in sequential recommendation. On the other hand, ITPS-BERT4Rec shows the best performance overall in approximating the Oracle. These findings validate the robustness of the proposed ITPS estimator in handling even extreme levels of exposure bias, and in capturing the true relevance of the items in a sequence and temporal context. 
Finally, as opposed to IPS-BERT4Rec which shows a significantly high and increasing variance, ITPS-BERT4Rec shows a relatively low and steady variance that compares to the variance of BERT4Rec. This further demonstrates the robustness of our proposed approach to increasing levels of exposure bias.

\subsubsection{RQ3: How important is an unbiased evaluation in assessing exposure debiasing?}

\begin{table}
\caption{Average R@k and NDCG@k over 5 replicates obtained with a standard evaluation process. $\uparrow$ means the ranking increased and $\downarrow$ means the ranking decreased compared to the unbiased results from section \ref{sec:rq1_results}. Best results are in \textbf{bold} and second to best are \underline{underlined}. A value with * is significantly higher than the next best value (p-value $< 0.05$).\label{tab:biased_results} }
\begin{center}
{
\resizebox{\columnwidth}{!}{\begin{tabular}{ l | c c c c}
\hline
Model & R@10 & NDCG@10 & R@5 & NDCG@5\\ \hline \hline
BERT4Rec & 0.7782 $\downarrow$ & 0.5851 $\downarrow$ & 0.6655 $\downarrow$ & 0.5486\\
IPS-BERT4Rec & 0.7835 $\uparrow$ & 0.5854 $\uparrow$ & 0.6665 $\uparrow$ & 0.5475\\
ITPS-BERT4Rec & \underline{0.7873}* & \underline{0.5909}* & \underline{0.6754}* & \underline{0.5545}\\
Oracle & \textbf{0.8000} & \textbf{0.5983} & \textbf{0.6795} & \textbf{0.5593}\\
\hline
\end{tabular}}
}
\end{center}
\end{table}

In this research question, we aim to demonstrate the importance of the unbiased evaluation process, explained in Section~\ref{sec:evaluation_process}, in evaluating the capacity of the models in capturing the true preferences of the users. To do so, we try to re-evaluate the tuned models using a standard Leave One Out (LOO) evaluation process, in which we compare the interacted test items to 100 randomly sampled items. This evaluation process is biased because the test items are not necessarily the most relevant items due to their exposure requirement. 
This results in an overestimation of the performance of the biased models, and their capacity to capture the true relevance. We summarize the results obtained with the standard LOO evaluation process in Table~\ref{tab:biased_results}.
We notice a discrepancy between the results obtained with the standard and unbiased evaluation processes. In fact, with the standard evaluation process, IPS-BERT4Rec outperformed BERT4Rec in almost all the settings, which reflects an over-estimation of the debiasing capabilities of the IPS framework and its ability to capture the relevance of items given the sequence context. The ITPS-BERT4Rec model was nonetheless still the top performer following the Oracle. These findings validate the necessity of relying on the unbiased evaluation setting, as it allows us to truly evaluate the properties of the different estimators.

\subsection{Experiments on Real Data}\label{sec:real_experiments}

We perform offline experiments on real recommendation datasets that aim to answer the following research questions:

\textbf{RQ4: }How well does our proposed ITPS estimator perform in terms of ranking accuracy?

\textbf{RQ5: }How well does our proposed ITPS estimator help mitigate popularity bias in the short and long terms?


\subsubsection{Data}

We rely on three datasets that are commonly used in sequential recommendation research \cite{sun2019bert4rec}, which are: the Movielens 1M (ml-1m)\footnote{\label{movielens_dataset_link}https://grouplens.org/datasets/movielens} \cite{10.1145/2827872}, Movielens 20M (ml-20m)\footnotemark[\getrefnumber{movielens_dataset_link}] \cite{10.1145/2827872}, and Amazon Beauty (beauty)\footnote{https://nijianmo.github.io/amazon/index.html} \cite{mcauley2015image}. For each of the datasets, we consider any rating, regardless of its value, as a positive interaction, then, we filter out users with less than 5 interactions to reduce the data sparsity. The dataset statistics are summarized in Table~\ref{tab:real_data_statistics}.

\begin{table}
\caption{Real dataset statistics.\label{tab:real_data_statistics} }
\begin{center}
{
\resizebox{\columnwidth}{!}{\begin{tabular}{ l | c c c c c c}
\hline
Dataset & Task & Sequences & Items & Interactions & Avg. length & Sparsity\\ \hline \hline
ml-1m & Movie rec. & 6,040 & 3,416 & 999,611 & 165.49 & 95.15\%\\
ml-20m & Movie rec. & 138,493 & 18,345 & 19,984,024 & 144.29 & 99.21\%\\
beauty & Product rec. & 40,226 & 54,542 & 353,962 & 8.79 & 99.98\%\\
\hline
\end{tabular}}
}
\end{center}
\end{table}

\subsubsection{Evaluation and Propensity Estimation}\label{sec:eval_prop_estimation}

Previously (Section~\ref{sec:semi_synthetic_experiments}), we were able to train our models using the true (temporal) exposure propensities and evaluate their ability to model the relevance using the temporal relevance levels, which were available through the use of semi-synthetic data. However, in real-world data, neither the (temporal) exposure propensities, nor the temporal relevance levels are available. This causes the following two issues: (1) We cannot evaluate the models' ability to learn the true relevance of the items to the users because we do not know the true temporal relevance levels; and (2) we cannot train the IPS-BERT4Rec and ITPS-BERT4Rec models as they rely on the exposure and temporal exposure propensities.
To solve the first issue, we propose an evaluation process that is based on popularity-based negative sampling. In fact, the main issue with the standard LOO evaluation process is that some of the randomly sampled negative items to which we are comparing our test and validation items may be as relevant, or possibly more relevant, than the test and validation items. We propose to sample the negative items for every sequence based on their popularities, meaning the higher the popularity of an item, the higher the probability that it will be sampled as a negative item. The idea is that more popular items have a higher likelihood that they have been exposed to the user and have not been interacted with because of their irrelevance to the user. The latter popularity-based negative sampling does not completely eliminate exposure bias in the evaluation. However, it is intended to mitigate it. Note that using popularity-based sampling to mitigate exposure bias was used in previous work \cite{gantner2012personalized} in the training phase. We are extending it to evaluation.
To solve the second issue, we build on previous work \cite{saito2019unbiased, damak2021debiased} and estimate the temporal exposure propensity of an item to a user by the temporal popularity of the item such that:

\begin{equation}
    \hat{\theta}_{S_{s}, I_{i}, t} = \frac{\sum_{j = 1}^{|S|} Y_{S_{j}, I_{i}, t}}{\sum_{k = 1}^{T}\sum_{l = 1}^{|I|}\sum_{j = 1}^{|S|} Y_{S_{j}, I_{l}, k}}.
\end{equation}

Similarly, we estimate the static exposure propensity of an item in a sequence with the item's popularity, which corresponds to the sum of the estimated temporal exposure propensities expressed as follows: $\hat{\theta}_{S_{s}, I_{i}} = \sum_{t = 1}^{T} \hat{\theta}_{S_{s}, I_{i}, t}.$
Thus, we train the IPS-BERT4Rec and ITPS-BERT4Rec models, presented in section~\ref{sec:models_compared}, using the estimated exposure propensities and estimated temporal exposure propensities, respectively.

\subsubsection{Hyperparameter Tuning}

For the beauty and ml-1m datasets, we perform the same hyperparameter tuning process described in Section~\ref{sec:tuning_semi_synthetic} on the semi-synthetic dataset. However, for the ml-20m dataset, we increase the ranges of some of the hyperparameters given the relatively higher size and complexity of the dataset. 
Hence, the number of hidden units is tuned within \{64, 128, 256\}, the number of transformer blocks within \{1, 2, 3\}, the number of attention heads within \{1, 2, 4, 8\}, the batch size within \{64, 128, 256\}, and the dropout rate within \{0, 0.01, 0.1, 0.2\}.

\subsubsection{RQ4: How well does the proposed ITPS estimator perform in terms of ranking accuracy?}

To measure the ranking capabilities of the proposed approach, we evaluate the tuned models using the evaluation process presented in Section~\ref{sec:eval_prop_estimation} which ensures that exposure bias is mitigated. Thus, the ranking accuracy results should provide a good approximation of how well the models capture the true relevance of the items to the users. We summarize the results on the three datasets in Table~\ref{tab:real_ranking_results}.
Our proposed ITPS-BERT4Rec model was the best performer in all the settings, showing significantly superior performance than the BERT4Rec and the IPS-BERT4Rec models in all the metrics and on all the datasets. This validates the ability of the proposed ITPS debiasing framework to learn the true relevance of the items to the users, in addition to its applicability in real recommendation settings. Moreover, interestingly, the ranking performance was not consistent for the second to best model. In fact, IPS-BERT4Rec outperformed BERT4Rec overall on both the ml-1m and beauty datasets but not on the ml-20m dataset.

\begin{table*}
\caption{Average Recall (R) and NDCG (N) results over 5 replicates on the three real interaction datasets. The best results are in \textbf{bold} and second to best results are \underline{underlined}. A value with * is significantly higher than the next best value (p-value $< 0.05$). \label{tab:real_ranking_results} }
\begin{center}
{
\resizebox{\textwidth}{!}{\begin{tabular}{ l | c c c c | c c c c | c c c c}
\hline
Dataset & \multicolumn{4}{c|}{ml-1m} & \multicolumn{4}{c|}{ml-20m} & \multicolumn{4}{c}{beauty}\\ \hline
Model & N@5 & R@5 & N@10 & R@10 & N@5 & R@5 & N@10 & R@10 & N@5 & R@5 & N@10 & R@10\\ \hline \hline
BERT4Rec & 0.2820 & 0.4086 & 0.3262 & 0.5454 & \underline{0.4205}* & \underline{0.5583}* & \underline{0.4624}* & \underline{0.6876}* & \underline{0.1056} & 0.1516 & 0.1260 & 0.2148\\
IPS-BERT4Rec & \underline{0.3416}* & \underline{0.4751}* & \underline{0.3801}* & \underline{0.5940}* & 0.4004 & 0.5389 & 0.4434 & 0.6715 & 0.1053 & \underline{0.1528} & \underline{0.1268} & \underline{0.2195}\\
ITPS-BERT4Rec & \textbf{0.3451}* & \textbf{0.4796} & \textbf{0.3844}* & \textbf{0.6007}* & \textbf{0.4295}* & \textbf{0.5674}* & \textbf{0.4709}* & \textbf{0.6952}* & \textbf{0.1197}* & \textbf{0.1745}* & \textbf{0.1444}* & \textbf{0.2510}*\\
\hline
\end{tabular}}
}
\end{center}
\end{table*}

\subsubsection{RQ5: How well does the proposed ITPS estimator help mitigate popularity bias in the short and long terms?}

To answer this question, we implement a feedback loop which simulates a real recommendation environment. The feedback loop consists of consecutive recommendation iterations where at each iteration, the recommender system is re-trained and generates top 10 recommendations for every user in the dataset. Each user then interacts with one of the recommended items and the interactions are added to the dataset for training future iterations. We simulate the user's choice with a uniform distribution, meaning that the interacted item is chosen at random from the recommendation list. Moreover, the choice of re-training the model at each iteration is related to the nature of our training datasets. In fact, we assume that an iteration corresponds to one day and that users interact with at most one movie or beauty product per day. This setting could be extended to other types of recommendation datasets in the future. Finally, we assume that all the users interact with one item at every iteration. As was discussed in \cite{ferraro2020exploring}, this assumption is meant to speed-up the feedback loop process and should not alter the general characteristics of the emerging phenomena. Thus, no conclusions will be altered.
We evaluate the popularity debiasing capabilities by looking at the novelty of the top 10 recommendations. The novelty is assessed using the Expected Free Discovery (EFD) \cite{vargas2011rank}, which is a measure of the ability of a system to recommend relevant long-tail items \cite{vargas2011rank} and is calculated as follows

\begin{equation}
    EFD@\mathcal{K}(TopK) = - \frac{1}{|S|} \sum_{s=1}^{|S|} \frac{1}{\mathcal{K}} \sum_{i \in TopK(S_{s})} log_{2} \hat{\theta}_{S_{s}, i},
\end{equation}

\noindent
where $TopK$ is the top $\mathcal{K}$ recommendation matrix in which every row represents the Top $\mathcal{K}$ recommendations in a sequence.

\begin{figure}
\begin{center}
\includegraphics[width=0.325\linewidth]{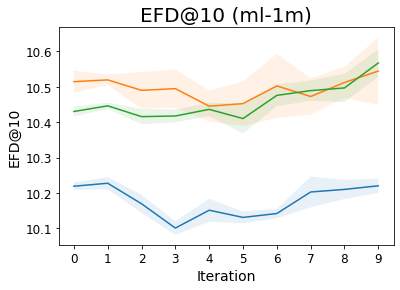}
\includegraphics[width=0.325\linewidth]{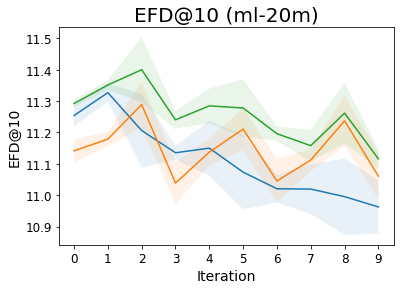}
\includegraphics[width=0.325\linewidth]{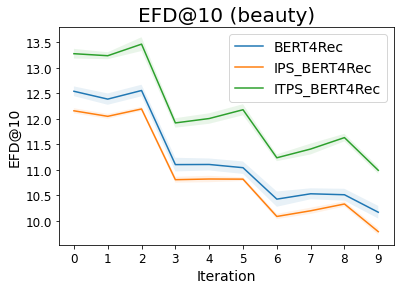}
\end{center}
\caption{Evolution of EFD@10 with respect to feedback loop iterations. All values are averages over 5 replicates and 90\% confidence intervals are highlighted. ITPS-BERT4Rec showed the best short and long-term popularity debiasing capabilities on the ml-20m and beauty datasets. \label{fig:feedback_loop_results}}
\end{figure}

We summarize the evolution of $EFD@10$ for 10 feedback loop iterations on the three datasets in Figure~\ref{fig:feedback_loop_results}. On both the ml-20m and beauty datasets, our proposed ITPS-BERT4Rec model showed the best results in all iterations. The difference in performance compared to the other two models was significant in all the iterations for the beauty dataset and in most iterations for the ml-20m dataset. However, we notice a change in trend in the ml-1m dataset where IPS-BERT4Rec and ITPS-BERT4Rec showed a relatively similar popularity debiasing performance, that still outperformed BERT4Rec. We believe that the difference in trend in the ml-1m dataset is due to the relatively low number of items and low sparsity of the dataset making the popularity bias problem less prominent compared to the other datasets. Moreover and interestingly, the vanilla BERT4Rec outperformed IPS-BERT4Rec on the beauty dataset. The overall superior performance of our proposed ITPS-BERT4Rec model shows the impact of exposure debiasing on popularity debiasing, where modeling the true preferences of the user results in more diverse and novel recommendations yielding a higher item discovery by the user.
Moreover, the ml-20m and beauty datasets showed, overall, decreasing trends for $EFD$ with respect to the feedback loop iterations for all the models. This means that the issue of popularity bias tends to worsen with time. However, the relatively low slope of ITPS-BERT4Rec demonstrates the importance of mitigating exposure bias to mitigate long-term popularity bias.

\section{Conclusion}

We studied the problem of exposure bias in sequential recommendation within the scope of bidirectional transformers trained to optimize the Cloze task, and proposed an ideal Cloze task loss that captures the true relevance. Then, we argued and proved that IPS estimators do not extend to sequential recommendation. In addition, we proposed a theoretically unbiased estimator for the ideal Cloze task loss, and formulated a framework that allows for an unbiased training and evaluation of sequential recommender systems. Our experiments empirically validated our claims of debiasing of the proposed ITPS-BERT4Rec estimator, and demonstrated its robustness to increasing levels of exposure bias, along with its longitudinal impact on popularity debiasing. Future work should validate and challenge the assumptions on which our theory is based.

\begin{acks}
This work was supported in part by National Science Foundation grants IIS-1549981, DRL-2026584, and CNS-1828521.
\end{acks}

\bibliographystyle{ACM-Reference-Format}
\bibliography{sample-base}


\begin{thebibliography}{41}


\ifx \showCODEN    \undefined \def \showCODEN     #1{\unskip}     \fi
\ifx \showDOI      \undefined \def \showDOI       #1{#1}\fi
\ifx \showISBNx    \undefined \def \showISBNx     #1{\unskip}     \fi
\ifx \showISBNxiii \undefined \def \showISBNxiii  #1{\unskip}     \fi
\ifx \showISSN     \undefined \def \showISSN      #1{\unskip}     \fi
\ifx \showLCCN     \undefined \def \showLCCN      #1{\unskip}     \fi
\ifx \shownote     \undefined \def \shownote      #1{#1}          \fi
\ifx \showarticletitle \undefined \def \showarticletitle #1{#1}   \fi
\ifx \showURL      \undefined \def \showURL       {\relax}        \fi
\providecommand\bibfield[2]{#2}
\providecommand\bibinfo[2]{#2}
\providecommand\natexlab[1]{#1}
\providecommand\showeprint[2][]{arXiv:#2}

\bibitem[Adomavicius et~al\mbox{.}(2005)]%
        {adomavicius2005incorporating}
\bibfield{author}{\bibinfo{person}{Gediminas Adomavicius},
  \bibinfo{person}{Ramesh Sankaranarayanan}, \bibinfo{person}{Shahana Sen},
  {and} \bibinfo{person}{Alexander Tuzhilin}.} \bibinfo{year}{2005}\natexlab{}.
\newblock \showarticletitle{Incorporating contextual information in recommender
  systems using a multidimensional approach}.
\newblock \bibinfo{journal}{\emph{ACM Transactions on Information Systems
  (TOIS)}} \bibinfo{volume}{23}, \bibinfo{number}{1} (\bibinfo{year}{2005}),
  \bibinfo{pages}{103--145}.
\newblock


\bibitem[Chen et~al\mbox{.}(2020a)]%
        {chen2020bias}
\bibfield{author}{\bibinfo{person}{Jiawei Chen}, \bibinfo{person}{Hande Dong},
  \bibinfo{person}{Xiang Wang}, \bibinfo{person}{Fuli Feng},
  \bibinfo{person}{Meng Wang}, {and} \bibinfo{person}{Xiangnan He}.}
  \bibinfo{year}{2020}\natexlab{a}.
\newblock \showarticletitle{Bias and Debias in Recommender System: A Survey and
  Future Directions}.
\newblock \bibinfo{journal}{\emph{arXiv preprint arXiv:2010.03240}}
  (\bibinfo{year}{2020}).
\newblock


\bibitem[Chen et~al\mbox{.}(2020b)]%
        {chen2020fast}
\bibfield{author}{\bibinfo{person}{Jiawei Chen}, \bibinfo{person}{Can Wang},
  \bibinfo{person}{Sheng Zhou}, \bibinfo{person}{Qihao Shi},
  \bibinfo{person}{Jingbang Chen}, \bibinfo{person}{Yan Feng}, {and}
  \bibinfo{person}{Chun Chen}.} \bibinfo{year}{2020}\natexlab{b}.
\newblock \showarticletitle{Fast Adaptively Weighted Matrix Factorization for
  Recommendation with Implicit Feedback.}. In \bibinfo{booktitle}{\emph{AAAI}}.
  \bibinfo{pages}{3470--3477}.
\newblock


\bibitem[Chen et~al\mbox{.}(2019)]%
        {chen2019samwalker}
\bibfield{author}{\bibinfo{person}{Jiawei Chen}, \bibinfo{person}{Can Wang},
  \bibinfo{person}{Sheng Zhou}, \bibinfo{person}{Qihao Shi},
  \bibinfo{person}{Yan Feng}, {and} \bibinfo{person}{Chun Chen}.}
  \bibinfo{year}{2019}\natexlab{}.
\newblock \showarticletitle{Samwalker: Social recommendation with informative
  sampling strategy}. In \bibinfo{booktitle}{\emph{The World Wide Web
  Conference}}. \bibinfo{pages}{228--239}.
\newblock


\bibitem[Cho et~al\mbox{.}(2014)]%
        {D14-1179}
\bibfield{author}{\bibinfo{person}{Kyunghyun Cho}, \bibinfo{person}{Bart van
  Merrienboer}, \bibinfo{person}{Caglar Gulcehre}, \bibinfo{person}{Dzmitry
  Bahdanau}, \bibinfo{person}{Fethi Bougares}, \bibinfo{person}{Holger
  Schwenk}, {and} \bibinfo{person}{Yoshua Bengio}.}
  \bibinfo{year}{2014}\natexlab{}.
\newblock \showarticletitle{Learning Phrase Representations using RNN
  Encoder--Decoder for Statistical Machine Translation}. In
  \bibinfo{booktitle}{\emph{Proceedings of the 2014 Conference on Empirical
  Methods in Natural Language Processing (EMNLP)}} (Doha, Qatar).
  \bibinfo{publisher}{Association for Computational Linguistics},
  \bibinfo{pages}{1724--1734}.
\newblock
\urldef\tempurl%
\url{https://doi.org/10.3115/v1/D14-1179}
\showDOI{\tempurl}


\bibitem[Damak et~al\mbox{.}(2021)]%
        {damak2021debiased}
\bibfield{author}{\bibinfo{person}{Khalil Damak}, \bibinfo{person}{Sami
  Khenissi}, {and} \bibinfo{person}{Olfa Nasraoui}.}
  \bibinfo{year}{2021}\natexlab{}.
\newblock \showarticletitle{Debiased Explainable Pairwise Ranking from Implicit
  Feedback}. In \bibinfo{booktitle}{\emph{Fifteenth ACM Conference on
  Recommender Systems}}. \bibinfo{pages}{321--331}.
\newblock


\bibitem[Devlin et~al\mbox{.}(2018)]%
        {devlin2018bert}
\bibfield{author}{\bibinfo{person}{Jacob Devlin}, \bibinfo{person}{Ming-Wei
  Chang}, \bibinfo{person}{Kenton Lee}, {and} \bibinfo{person}{Kristina
  Toutanova}.} \bibinfo{year}{2018}\natexlab{}.
\newblock \showarticletitle{Bert: Pre-training of deep bidirectional
  transformers for language understanding}.
\newblock \bibinfo{journal}{\emph{arXiv preprint arXiv:1810.04805}}
  (\bibinfo{year}{2018}).
\newblock


\bibitem[Devooght et~al\mbox{.}(2015)]%
        {devooght2015dynamic}
\bibfield{author}{\bibinfo{person}{Robin Devooght}, \bibinfo{person}{Nicolas
  Kourtellis}, {and} \bibinfo{person}{Amin Mantrach}.}
  \bibinfo{year}{2015}\natexlab{}.
\newblock \showarticletitle{Dynamic matrix factorization with priors on unknown
  values}. In \bibinfo{booktitle}{\emph{Proceedings of the 21th ACM SIGKDD
  international conference on knowledge discovery and data mining}}.
  \bibinfo{pages}{189--198}.
\newblock


\bibitem[Ferraro et~al\mbox{.}(2020)]%
        {ferraro2020exploring}
\bibfield{author}{\bibinfo{person}{Andres Ferraro}, \bibinfo{person}{Dietmar
  Jannach}, {and} \bibinfo{person}{Xavier Serra}.}
  \bibinfo{year}{2020}\natexlab{}.
\newblock \showarticletitle{Exploring Longitudinal Effects of Session-based
  Recommendations}. In \bibinfo{booktitle}{\emph{Fourteenth ACM Conference on
  Recommender Systems}}. \bibinfo{pages}{474--479}.
\newblock


\bibitem[Gantner et~al\mbox{.}(2012)]%
        {gantner2012personalized}
\bibfield{author}{\bibinfo{person}{Zeno Gantner}, \bibinfo{person}{Lucas
  Drumond}, \bibinfo{person}{Christoph Freudenthaler}, {and}
  \bibinfo{person}{Lars Schmidt-Thieme}.} \bibinfo{year}{2012}\natexlab{}.
\newblock \showarticletitle{Personalized ranking for non-uniformly sampled
  items}. In \bibinfo{booktitle}{\emph{Proceedings of KDD Cup 2011}}. PMLR,
  \bibinfo{pages}{231--247}.
\newblock


\bibitem[Harper and Konstan(2015)]%
        {10.1145/2827872}
\bibfield{author}{\bibinfo{person}{F.~Maxwell Harper} {and}
  \bibinfo{person}{Joseph~A. Konstan}.} \bibinfo{year}{2015}\natexlab{}.
\newblock \showarticletitle{The MovieLens Datasets: History and Context}.
\newblock \bibinfo{journal}{\emph{ACM Trans. Interact. Intell. Syst.}}
  \bibinfo{volume}{5}, \bibinfo{number}{4}, Article \bibinfo{articleno}{19}
  (\bibinfo{date}{dec} \bibinfo{year}{2015}), \bibinfo{numpages}{19}~pages.
\newblock
\showISSN{2160-6455}
\urldef\tempurl%
\url{https://doi.org/10.1145/2827872}
\showDOI{\tempurl}


\bibitem[He et~al\mbox{.}(2016)]%
        {he2016fast}
\bibfield{author}{\bibinfo{person}{Xiangnan He}, \bibinfo{person}{Hanwang
  Zhang}, \bibinfo{person}{Min-Yen Kan}, {and} \bibinfo{person}{Tat-Seng
  Chua}.} \bibinfo{year}{2016}\natexlab{}.
\newblock \showarticletitle{Fast matrix factorization for online recommendation
  with implicit feedback}. In \bibinfo{booktitle}{\emph{Proceedings of the 39th
  International ACM SIGIR conference on Research and Development in Information
  Retrieval}}. \bibinfo{pages}{549--558}.
\newblock


\bibitem[Hidasi and Karatzoglou(2018)]%
        {hidasi2018recurrent}
\bibfield{author}{\bibinfo{person}{Bal{\'a}zs Hidasi} {and}
  \bibinfo{person}{Alexandros Karatzoglou}.} \bibinfo{year}{2018}\natexlab{}.
\newblock \showarticletitle{Recurrent neural networks with top-k gains for
  session-based recommendations}. In \bibinfo{booktitle}{\emph{Proceedings of
  the 27th ACM international conference on information and knowledge
  management}}. \bibinfo{pages}{843--852}.
\newblock


\bibitem[Hidasi et~al\mbox{.}(2015)]%
        {hidasi2015session}
\bibfield{author}{\bibinfo{person}{Bal{\'a}zs Hidasi},
  \bibinfo{person}{Alexandros Karatzoglou}, \bibinfo{person}{Linas Baltrunas},
  {and} \bibinfo{person}{Domonkos Tikk}.} \bibinfo{year}{2015}\natexlab{}.
\newblock \showarticletitle{Session-based recommendations with recurrent neural
  networks}.
\newblock \bibinfo{journal}{\emph{arXiv preprint arXiv:1511.06939}}
  (\bibinfo{year}{2015}).
\newblock


\bibitem[Hochreiter and Schmidhuber(1997)]%
        {Hochreiter:1997:LSM:1246443.1246450}
\bibfield{author}{\bibinfo{person}{Sepp Hochreiter} {and}
  \bibinfo{person}{J\"{u}rgen Schmidhuber}.} \bibinfo{year}{1997}\natexlab{}.
\newblock \showarticletitle{Long Short-Term Memory}.
\newblock \bibinfo{journal}{\emph{Neural Comput.}} \bibinfo{volume}{9},
  \bibinfo{number}{8} (\bibinfo{date}{Nov.} \bibinfo{year}{1997}),
  \bibinfo{pages}{1735--1780}.
\newblock
\showISSN{0899-7667}
\urldef\tempurl%
\url{https://doi.org/10.1162/neco.1997.9.8.1735}
\showDOI{\tempurl}


\bibitem[Hu et~al\mbox{.}(2008)]%
        {hu2008collaborative}
\bibfield{author}{\bibinfo{person}{Yifan Hu}, \bibinfo{person}{Yehuda Koren},
  {and} \bibinfo{person}{Chris Volinsky}.} \bibinfo{year}{2008}\natexlab{}.
\newblock \showarticletitle{Collaborative filtering for implicit feedback
  datasets}. In \bibinfo{booktitle}{\emph{2008 Eighth IEEE International
  Conference on Data Mining}}. Ieee, \bibinfo{pages}{263--272}.
\newblock


\bibitem[Kang and McAuley(2018)]%
        {kang2018self}
\bibfield{author}{\bibinfo{person}{Wang-Cheng Kang} {and}
  \bibinfo{person}{Julian McAuley}.} \bibinfo{year}{2018}\natexlab{}.
\newblock \showarticletitle{Self-attentive sequential recommendation}. In
  \bibinfo{booktitle}{\emph{2018 IEEE International Conference on Data Mining
  (ICDM)}}. IEEE, \bibinfo{pages}{197--206}.
\newblock


\bibitem[LeCun et~al\mbox{.}(1999)]%
        {lecun1999object}
\bibfield{author}{\bibinfo{person}{Yann LeCun}, \bibinfo{person}{Patrick
  Haffner}, \bibinfo{person}{L{\'e}on Bottou}, {and} \bibinfo{person}{Yoshua
  Bengio}.} \bibinfo{year}{1999}\natexlab{}.
\newblock \showarticletitle{Object recognition with gradient-based learning}.
\newblock In \bibinfo{booktitle}{\emph{Shape, contour and grouping in computer
  vision}}. \bibinfo{publisher}{Springer}, \bibinfo{pages}{319--345}.
\newblock


\bibitem[Li et~al\mbox{.}(2010)]%
        {li2010improving}
\bibfield{author}{\bibinfo{person}{Yanen Li}, \bibinfo{person}{Jia Hu},
  \bibinfo{person}{ChengXiang Zhai}, {and} \bibinfo{person}{Ye Chen}.}
  \bibinfo{year}{2010}\natexlab{}.
\newblock \showarticletitle{Improving one-class collaborative filtering by
  incorporating rich user information}. In
  \bibinfo{booktitle}{\emph{Proceedings of the 19th ACM international
  conference on Information and knowledge management}}.
  \bibinfo{pages}{959--968}.
\newblock


\bibitem[Liang et~al\mbox{.}(2016)]%
        {liang2016modeling}
\bibfield{author}{\bibinfo{person}{Dawen Liang}, \bibinfo{person}{Laurent
  Charlin}, \bibinfo{person}{James McInerney}, {and} \bibinfo{person}{David~M
  Blei}.} \bibinfo{year}{2016}\natexlab{}.
\newblock \showarticletitle{Modeling user exposure in recommendation}. In
  \bibinfo{booktitle}{\emph{Proceedings of the 25th international conference on
  World Wide Web}}. \bibinfo{pages}{951--961}.
\newblock


\bibitem[Lipton(2015)]%
        {Lipton2015ACR}
\bibfield{author}{\bibinfo{person}{Zachary~Chase Lipton}.}
  \bibinfo{year}{2015}\natexlab{}.
\newblock \showarticletitle{A Critical Review of Recurrent Neural Networks for
  Sequence Learning}.
\newblock \bibinfo{journal}{\emph{CoRR}}  \bibinfo{volume}{abs/1506.00019}
  (\bibinfo{year}{2015}).
\newblock


\bibitem[McAuley et~al\mbox{.}(2015)]%
        {mcauley2015image}
\bibfield{author}{\bibinfo{person}{Julian McAuley},
  \bibinfo{person}{Christopher Targett}, \bibinfo{person}{Qinfeng Shi}, {and}
  \bibinfo{person}{Anton Van Den~Hengel}.} \bibinfo{year}{2015}\natexlab{}.
\newblock \showarticletitle{Image-based recommendations on styles and
  substitutes}. In \bibinfo{booktitle}{\emph{Proceedings of the 38th
  international ACM SIGIR conference on research and development in information
  retrieval}}. \bibinfo{pages}{43--52}.
\newblock


\bibitem[Pan and Scholz(2009)]%
        {pan2009mind}
\bibfield{author}{\bibinfo{person}{Rong Pan} {and} \bibinfo{person}{Martin
  Scholz}.} \bibinfo{year}{2009}\natexlab{}.
\newblock \showarticletitle{Mind the gaps: weighting the unknown in large-scale
  one-class collaborative filtering}. In \bibinfo{booktitle}{\emph{Proceedings
  of the 15th ACM SIGKDD international conference on Knowledge discovery and
  data mining}}. \bibinfo{pages}{667--676}.
\newblock


\bibitem[Pan et~al\mbox{.}(2008)]%
        {pan2008one}
\bibfield{author}{\bibinfo{person}{Rong Pan}, \bibinfo{person}{Yunhong Zhou},
  \bibinfo{person}{Bin Cao}, \bibinfo{person}{Nathan~N Liu},
  \bibinfo{person}{Rajan Lukose}, \bibinfo{person}{Martin Scholz}, {and}
  \bibinfo{person}{Qiang Yang}.} \bibinfo{year}{2008}\natexlab{}.
\newblock \showarticletitle{One-class collaborative filtering}. In
  \bibinfo{booktitle}{\emph{2008 Eighth IEEE International Conference on Data
  Mining}}. IEEE, \bibinfo{pages}{502--511}.
\newblock


\bibitem[Ranzato et~al\mbox{.}(2015)]%
        {ranzato2015sequence}
\bibfield{author}{\bibinfo{person}{Marc'Aurelio Ranzato},
  \bibinfo{person}{Sumit Chopra}, \bibinfo{person}{Michael Auli}, {and}
  \bibinfo{person}{Wojciech Zaremba}.} \bibinfo{year}{2015}\natexlab{}.
\newblock \showarticletitle{Sequence level training with recurrent neural
  networks}.
\newblock \bibinfo{journal}{\emph{arXiv preprint arXiv:1511.06732}}
  (\bibinfo{year}{2015}).
\newblock


\bibitem[Ren et~al\mbox{.}(2020)]%
        {ren2020sequential}
\bibfield{author}{\bibinfo{person}{Ruiyang Ren}, \bibinfo{person}{Zhaoyang
  Liu}, \bibinfo{person}{Yaliang Li}, \bibinfo{person}{Wayne~Xin Zhao},
  \bibinfo{person}{Hui Wang}, \bibinfo{person}{Bolin Ding}, {and}
  \bibinfo{person}{Ji-Rong Wen}.} \bibinfo{year}{2020}\natexlab{}.
\newblock \showarticletitle{Sequential recommendation with self-attentive
  multi-adversarial network}. In \bibinfo{booktitle}{\emph{Proceedings of the
  43rd International ACM SIGIR Conference on Research and Development in
  Information Retrieval}}. \bibinfo{pages}{89--98}.
\newblock


\bibitem[Saito(2019)]%
        {saito2019unbiased}
\bibfield{author}{\bibinfo{person}{Yuta Saito}.}
  \bibinfo{year}{2019}\natexlab{}.
\newblock \showarticletitle{Unbiased Pairwise Learning from Implicit Feedback}.
  In \bibinfo{booktitle}{\emph{NeurIPS 2019 Workshop on Causal Machine
  Learning}}.
\newblock


\bibitem[Saito et~al\mbox{.}(2020)]%
        {saito2020unbiased}
\bibfield{author}{\bibinfo{person}{Yuta Saito}, \bibinfo{person}{Suguru
  Yaginuma}, \bibinfo{person}{Yuta Nishino}, \bibinfo{person}{Hayato Sakata},
  {and} \bibinfo{person}{Kazuhide Nakata}.} \bibinfo{year}{2020}\natexlab{}.
\newblock \showarticletitle{Unbiased recommender learning from
  missing-not-at-random implicit feedback}. In
  \bibinfo{booktitle}{\emph{Proceedings of the 13th International Conference on
  Web Search and Data Mining}}. \bibinfo{pages}{501--509}.
\newblock


\bibitem[Schnabel et~al\mbox{.}(2016)]%
        {schnabel2016recommendations}
\bibfield{author}{\bibinfo{person}{Tobias Schnabel}, \bibinfo{person}{Adith
  Swaminathan}, \bibinfo{person}{Ashudeep Singh}, \bibinfo{person}{Navin
  Chandak}, {and} \bibinfo{person}{Thorsten Joachims}.}
  \bibinfo{year}{2016}\natexlab{}.
\newblock \showarticletitle{Recommendations as treatments: Debiasing learning
  and evaluation}. In \bibinfo{booktitle}{\emph{international conference on
  machine learning}}. PMLR, \bibinfo{pages}{1670--1679}.
\newblock


\bibitem[Sun et~al\mbox{.}(2019b)]%
        {sun2019bert4rec}
\bibfield{author}{\bibinfo{person}{Fei Sun}, \bibinfo{person}{Jun Liu},
  \bibinfo{person}{Jian Wu}, \bibinfo{person}{Changhua Pei},
  \bibinfo{person}{Xiao Lin}, \bibinfo{person}{Wenwu Ou}, {and}
  \bibinfo{person}{Peng Jiang}.} \bibinfo{year}{2019}\natexlab{b}.
\newblock \showarticletitle{BERT4Rec: Sequential recommendation with
  bidirectional encoder representations from transformer}. In
  \bibinfo{booktitle}{\emph{Proceedings of the 28th ACM international
  conference on information and knowledge management}}.
  \bibinfo{pages}{1441--1450}.
\newblock


\bibitem[Sun et~al\mbox{.}(2019a)]%
        {sun2019debiasing}
\bibfield{author}{\bibinfo{person}{Wenlong Sun}, \bibinfo{person}{Sami
  Khenissi}, \bibinfo{person}{Olfa Nasraoui}, {and} \bibinfo{person}{Patrick
  Shafto}.} \bibinfo{year}{2019}\natexlab{a}.
\newblock \showarticletitle{Debiasing the human-recommender system feedback
  loop in collaborative filtering}. In \bibinfo{booktitle}{\emph{Companion
  Proceedings of The 2019 World Wide Web Conference}}.
  \bibinfo{pages}{645--651}.
\newblock


\bibitem[Tang and Wang(2018)]%
        {tang2018personalized}
\bibfield{author}{\bibinfo{person}{Jiaxi Tang} {and} \bibinfo{person}{Ke
  Wang}.} \bibinfo{year}{2018}\natexlab{}.
\newblock \showarticletitle{Personalized top-n sequential recommendation via
  convolutional sequence embedding}. In \bibinfo{booktitle}{\emph{Proceedings
  of the Eleventh ACM International Conference on Web Search and Data Mining}}.
  \bibinfo{pages}{565--573}.
\newblock


\bibitem[Taylor(1953)]%
        {taylor1953cloze}
\bibfield{author}{\bibinfo{person}{Wilson~L Taylor}.}
  \bibinfo{year}{1953}\natexlab{}.
\newblock \showarticletitle{“Cloze procedure”: A new tool for measuring
  readability}.
\newblock \bibinfo{journal}{\emph{Journalism quarterly}} \bibinfo{volume}{30},
  \bibinfo{number}{4} (\bibinfo{year}{1953}), \bibinfo{pages}{415--433}.
\newblock


\bibitem[Vargas and Castells(2011)]%
        {vargas2011rank}
\bibfield{author}{\bibinfo{person}{Sa{\'u}l Vargas} {and}
  \bibinfo{person}{Pablo Castells}.} \bibinfo{year}{2011}\natexlab{}.
\newblock \showarticletitle{Rank and relevance in novelty and diversity metrics
  for recommender systems}. In \bibinfo{booktitle}{\emph{Proceedings of the
  fifth ACM conference on Recommender systems}}. \bibinfo{pages}{109--116}.
\newblock


\bibitem[Vaswani et~al\mbox{.}(2017)]%
        {vaswani2017attention}
\bibfield{author}{\bibinfo{person}{Ashish Vaswani}, \bibinfo{person}{Noam
  Shazeer}, \bibinfo{person}{Niki Parmar}, \bibinfo{person}{Jakob Uszkoreit},
  \bibinfo{person}{Llion Jones}, \bibinfo{person}{Aidan~N Gomez},
  \bibinfo{person}{Lukasz Kaiser}, {and} \bibinfo{person}{Illia Polosukhin}.}
  \bibinfo{year}{2017}\natexlab{}.
\newblock \showarticletitle{Attention is all you need}.
\newblock \bibinfo{journal}{\emph{arXiv preprint arXiv:1706.03762}}
  (\bibinfo{year}{2017}).
\newblock


\bibitem[Wang et~al\mbox{.}(2019)]%
        {wang2019survey}
\bibfield{author}{\bibinfo{person}{Shoujin Wang}, \bibinfo{person}{Longbing
  Cao}, \bibinfo{person}{Yan Wang}, \bibinfo{person}{Quan~Z Sheng},
  \bibinfo{person}{Mehmet Orgun}, {and} \bibinfo{person}{Defu Lian}.}
  \bibinfo{year}{2019}\natexlab{}.
\newblock \showarticletitle{A survey on session-based recommender systems}.
\newblock \bibinfo{journal}{\emph{arXiv preprint arXiv:1902.04864}}
  (\bibinfo{year}{2019}).
\newblock


\bibitem[Wang et~al\mbox{.}(2020)]%
        {wang2020reinforced}
\bibfield{author}{\bibinfo{person}{Xiang Wang}, \bibinfo{person}{Yaokun Xu},
  \bibinfo{person}{Xiangnan He}, \bibinfo{person}{Yixin Cao},
  \bibinfo{person}{Meng Wang}, {and} \bibinfo{person}{Tat-Seng Chua}.}
  \bibinfo{year}{2020}\natexlab{}.
\newblock \showarticletitle{Reinforced Negative Sampling over Knowledge Graph
  for Recommendation}. In \bibinfo{booktitle}{\emph{Proceedings of The Web
  Conference 2020}}. \bibinfo{pages}{99--109}.
\newblock


\bibitem[Yang et~al\mbox{.}(2018)]%
        {yang2018unbiased}
\bibfield{author}{\bibinfo{person}{Longqi Yang}, \bibinfo{person}{Yin Cui},
  \bibinfo{person}{Yuan Xuan}, \bibinfo{person}{Chenyang Wang},
  \bibinfo{person}{Serge Belongie}, {and} \bibinfo{person}{Deborah Estrin}.}
  \bibinfo{year}{2018}\natexlab{}.
\newblock \showarticletitle{Unbiased offline recommender evaluation for
  missing-not-at-random implicit feedback}. In
  \bibinfo{booktitle}{\emph{Proceedings of the 12th ACM Conference on
  Recommender Systems}}. \bibinfo{pages}{279--287}.
\newblock


\bibitem[Yu et~al\mbox{.}(2017)]%
        {yu2017selection}
\bibfield{author}{\bibinfo{person}{Hsiang-Fu Yu}, \bibinfo{person}{Mikhail
  Bilenko}, {and} \bibinfo{person}{Chih-Jen Lin}.}
  \bibinfo{year}{2017}\natexlab{}.
\newblock \showarticletitle{Selection of negative samples for one-class matrix
  factorization}. In \bibinfo{booktitle}{\emph{Proceedings of the 2017 SIAM
  International Conference on Data Mining}}. SIAM, \bibinfo{pages}{363--371}.
\newblock


\bibitem[Zhao et~al\mbox{.}(2021)]%
        {zhao2021tbtf}
\bibfield{author}{\bibinfo{person}{Jianli Zhao}, \bibinfo{person}{Shangcheng
  Yang}, \bibinfo{person}{Huan Huo}, \bibinfo{person}{Qiuxia Sun}, {and}
  \bibinfo{person}{Xijiao Geng}.} \bibinfo{year}{2021}\natexlab{}.
\newblock \showarticletitle{TBTF: an effective time-varying bias tensor
  factorization algorithm for recommender system}.
\newblock \bibinfo{journal}{\emph{Applied Intelligence}}
  (\bibinfo{year}{2021}), \bibinfo{pages}{1--12}.
\newblock


\bibitem[Zhao et~al\mbox{.}(2020)]%
        {zhao2020adversarial}
\bibfield{author}{\bibinfo{person}{Pengyu Zhao}, \bibinfo{person}{Tianxiao
  Shui}, \bibinfo{person}{Yuanxing Zhang}, \bibinfo{person}{Kecheng Xiao},
  {and} \bibinfo{person}{Kaigui Bian}.} \bibinfo{year}{2020}\natexlab{}.
\newblock \showarticletitle{Adversarial Oracular Seq2seq Learning for
  Sequential Recommendation}. In \bibinfo{booktitle}{\emph{Proceedings of the
  Twenty-Ninth International Joint Conference on Artificial Intelligence,
  IJCAI}}. \bibinfo{pages}{1905--1911}.
\newblock


\end{thebibliography}

\appendix

\section{Supplemental Material}

\subsection{Proof of Proposition \ref{prop:bias_cloze_loss}}\label{app:bias_cloze_loss}

\begin{proof}
\begin{align*}
&\mathbb{E}[L_{Cloze}] = \mathbb{E}[\frac{-1}{|S| |I| T} \sum_{s = 1}^{|S|} \sum_{t = 1}^{T} \sum_{i = 1}^{|I|} \mathds{1}_{\{ S^{m}_{s, t} = \langle mask \rangle \}} Y_{S_{s}, I_{i}, t}\\
    &\times log \; softmax(f_{\Omega}(S^{m}_{s, t}, I_{i}))]\\
&= \frac{-1}{|S| |I| T} \sum_{s = 1}^{|S|} \sum_{t = 1}^{T} \sum_{i = 1}^{|I|} \mathds{1}_{\{ S^{m}_{s, t} = \langle mask\rangle \}}\\
&\times C_{S_{s}, I_{i}, t} \; \theta_{S_{s}, I_{i}, t} \; \gamma_{S_{s}, I_{i}, t} log \; softmax(f_{\Omega}(S^{m}_{s, t}, I_{i}))
\end{align*}

Given that the temporal propensities $\theta_{S_{s}, I_{i}, t}$ cannot always be equal to $1$, $\forall (S_{s}, I_{i}, t) \in S \times I \times [1..T]$. Thus, $\mathbb{E}[L_{Cloze}] \neq L_{Cloze}^{ideal}$.
\end{proof}

Note that the proof relies on the probabilistic model of the interaction random variable that is  proposed later in Definition \ref{def:interaction_probabilistic_model}.

\subsection{Proof of Proposition \ref{prop:unbiasedness_condition}}\label{app:proof_unbiasedness_condition}

\begin{proof}
\begin{align*}
&\mathbb{E}[L_{Cloze}^{IPS}] = L_{Cloze}^{ideal}\\
&\Leftrightarrow \mathbb{E}[\frac{-1}{|S| |I| T} \sum_{s = 1}^{|S|} \sum_{t = 1}^{T} \sum_{i = 1}^{|I|} \mathds{1}_{\{ S^{m}_{s, t} = \langle mask \rangle \}} \frac{Y_{S_{s}, I_{i}, t}}{\theta_{S_{s}, I_{i}}}\\ 
    &\times log \; softmax(f_{\Omega}(S^{m}_{s, t}, I_{i}))]\\
    &= \frac{-1}{|S| |I| T} \sum_{s = 1}^{|S|} \sum_{t = 1}^{T} \sum_{i = 1}^{|I|} \mathds{1}_{\{ S^{m}_{s, t} = \langle mask \rangle \}} C_{S_{s}, I_{i}, t}\\ 
    &\times \gamma_{S_{s}, I_{i}, t} log \; softmax(f_{\Omega}(S^{m}_{s, t}, I_{i}))\\
&\Leftrightarrow \mathbb{E}[\frac{-1}{|S| |I| T} \sum_{s = 1}^{|S|} \sum_{t = 1}^{T} \sum_{i = 1}^{|I|} \mathds{1}_{\{ S^{m}_{s, t} = \langle mask \rangle \}} \frac{C_{S_{s}, I_{i}, t} \; O_{S_{s}, I_{i}, t} \; R_{S_{s}, I_{i}, t}}{\theta_{S_{s}, I_{i}}}\\ 
    &\times log \; softmax(f_{\Omega}(S^{m}_{s, t}, I_{i}))]\\
    &= \frac{-1}{|S| |I| T} \sum_{s = 1}^{|S|} \sum_{t = 1}^{T} \sum_{i = 1}^{|I|} \mathds{1}_{\{ S^{m}_{s, t} = \langle mask \rangle \}} C_{S_{s}, I_{i}, t}\\ 
    &\times \gamma_{S_{s}, I_{i}, t} log \; softmax(f_{\Omega}(S^{m}_{s, t}, I_{i}))\\
\end{align*}
\vfill\eject
\begin{align*}
&\Leftrightarrow \frac{-1}{|S| |I| T} \sum_{s = 1}^{|S|} \sum_{t = 1}^{T} \sum_{i = 1}^{|I|} \mathds{1}_{\{ S^{m}_{s, t} = \langle mask \rangle \}} \frac{C_{S_{s}, I_{i}, t} \; \theta_{S_{s}, I_{i}, t} \; \gamma_{S_{s}, I_{i}, t}}{\theta_{S_{s}, I_{i}}}\\ 
    &\times log \; softmax(f_{\Omega}(S^{m}_{s, t}, I_{i}))\\
    &= \frac{-1}{|S| |I| T} \sum_{s = 1}^{|S|} \sum_{t = 1}^{T} \sum_{i = 1}^{|I|} \mathds{1}_{\{ S^{m}_{s, t} = \langle mask \rangle \}} C_{S_{s}, I_{i}, t}\\ 
    &\times \gamma_{S_{s}, I_{i}, t} log \; softmax(f_{\Omega}(S^{m}_{s, t}, I_{i}))\\
&\Leftrightarrow \theta_{S_{s}, I_{i}, t} = \theta_{S_{s}, I_{i}}, \forall (S_{s}, I_{i}, t) \in S \times I \times [1..T].\qedhere
\end{align*}
\end{proof}

Note that the proof also relies on the probabilistic model of the interaction random variable that is  proposed later in Definition \ref{def:interaction_probabilistic_model}.

\subsection{Proof of Proposition \ref{prop:itps_unbiased}}\label{app:proof_itps_unbiased}

\begin{proof}
\begin{align*}
&\mathbb{E}[L_{Cloze}^{ITPS}] = \mathbb{E}[\frac{-1}{|S| |I| T} \sum_{s = 1}^{|S|} \sum_{t = 1}^{T} \sum_{i = 1}^{|I|} \mathds{1}_{\{ S^{m}_{s, t} = \langle mask \rangle \}}\\
&\times \frac{C_{S_{s}, I_{i}, t} \; O_{S_{s}, I_{i}, t} \; R_{S_{s}, I_{i}, t}}{\theta_{S_{s}, I_{i}, t}} log \; softmax(f_{\Omega}(S^{m}_{s, t}, I_{i}))]\\
&= \frac{-1}{|S| |I| T} \sum_{s = 1}^{|S|} \sum_{t = 1}^{T} \sum_{i = 1}^{|I|} \mathds{1}_{\{ S^{m}_{s, t} = \langle mask\rangle \}}\\
&\times \frac{C_{S_{s}, I_{i}, t} \; \theta_{S_{s}, I_{i}, t} \; \gamma_{S_{s}, I_{i}, t}}{\theta_{S_{s}, I_{i}, t}} log \; softmax(f_{\Omega}(S^{m}_{s, t}, I_{i}))\\
&= \frac{-1}{|S| |I| T} \sum_{s = 1}^{|S|} \sum_{t = 1}^{T} \sum_{i = 1}^{|I|} \mathds{1}_{\{ S^{m}_{s, t} = \langle mask \rangle \}} C_{S_{s}, I_{i}, t} \; \gamma_{S_{s}, I_{i}, t}\\
&\times log \; softmax(f_{\Omega}(S^{m}_{s, t}, I_{i})) = L_{Cloze}^{ideal}\qedhere
\end{align*}
\end{proof}

Note that the proof assumes independence between exposure and relevance. Also, it assumes that the outcome of the choice model for an item is deterministic, which is reasonable if we assume a rational user who tends to choose the most relevant item among the exposed items.

\end{document}